\documentclass[10pt,twocolumn,letterpaper]{article}

\usepackage[pagenumbers]{cvpr} 

\usepackage{graphicx}
\usepackage{amsmath}
\usepackage{amssymb}
\usepackage{booktabs}
\usepackage{multirow}
\usepackage[table]{xcolor}
\usepackage{arydshln}
\usepackage{hhline}
\usepackage[accsupp]{axessibility}

\usepackage[pagebackref,breaklinks,colorlinks]{hyperref}

\usepackage[capitalize]{cleveref}
\crefname{section}{Sec.}{Secs.}
\Crefname{section}{Section}{Sections}
\Crefname{table}{Table}{Tables}
\crefname{table}{Tab.}{Tabs.}

\def\cvprPaperID{12379} 
\def\confName{CVPR}
\def\confYear{2023}

\begin{document}

\title{FlexNeRF: Photorealistic Free-viewpoint Rendering of Moving Humans \\ from Sparse Views}

\author{Vinoj Jayasundara$^{1}$\thanks{Part of the work was done while the author was an intern at Amazon.}, Amit Agrawal$^{2}$, Nicolas Heron$^{2}$, Abhinav Shrivastava$^{1}$, Larry S. Davis$^{1,2}$\\
\\
$^1$University of Maryland, College Park \hspace{1em} $^2$Amazon.com, Inc.\\
{\tt\small \{vinoj, lsdavis\}@umd.edu, \{aaagrawa, heron\}@amazon.com, abhinav@cs.umd.edu}
}
\maketitle

\begin{abstract}
We present FlexNeRF, a method for photorealistic free-viewpoint rendering of humans in motion from monocular videos. Our approach works well with sparse views, which is a challenging scenario when the subject is exhibiting fast/complex motions. We propose a novel approach which jointly optimizes a canonical time and pose configuration, with a pose-dependent motion field and pose-independent temporal deformations complementing each other. Thanks to our novel temporal and cyclic consistency constraints along with additional losses on intermediate representation such as segmentation, our approach provides high quality outputs as the observed views become sparser. We empirically demonstrate that our method significantly outperforms the state-of-the-art on public benchmark datasets as well as a self-captured fashion dataset. The project page is available at: \url{https://flex-nerf.github.io/}.
\end{abstract}

\vspace{-5mm}
\section{Introduction}
\label{sec:intro}
Free-viewpoint rendering of a scene is an important problem often attempted under constrained settings: on subjects demonstrating simple motion carefully captured with multiple cameras \cite{Liu2020NeuralHV, MartinBrualla2018LookinGoodEP, MartinBrualla2021NeRFIT}. However, photorealistic \textit{free-viewpoint} rendering of moving humans captured from a \textit{monocular} video still remains an unsolved challenging problem, especially with sparse views.

Neural radiance fields (NeRF) have emerged as a popular tool to learn radiance fields from images/videos for novel view-point rendering. Previous approaches assume multiple view-points and often fail on non-rigid human motions. Human-specific NeRFs have recently become popular for learning models using input videos\cite{Weng2022HumanNeRFFR,Peng2021AnimatableNR}. The current state-of-art approaches such as HumanNeRF\cite{Weng2022HumanNeRFFR} have shown impressive progress in this domain. However, there remain several challenges. Firstly, approaches such as HumanNeRF\cite{Weng2022HumanNeRFFR} utilize a pose prior and use a canonical configuration (\eg T-pose) for optimization, which may be well outside the set of observed poses. The underlying optimization becomes challenging especially as the number of observed views become sparse. In contrast, we select a pose from the available set of poses as the canonical pose-configuration, similar to previous pose-free approaches such as D-NeRF\cite{Pumarola2021DNeRFNR}. This enables best of both worlds; it becomes easier to learn a motion field mapping due to smaller deformations while using a pose prior. In addition, having the canonical view in the training data provides a strong prior for the optimization of the canonical pose itself. Finally, it allows us to optimize the canonical configuration with our novel \textit{pose-independent temporal deformation}. We demonstrate that this architectural change provides significantly better results compared to existing approaches \cite{Weng2022HumanNeRFFR, Ouyang2022RealTimeNC}.

In addition, approaches such as HumanNeRF\cite{Weng2022HumanNeRFFR} depend on the estimated pose for the canonical configuration optimization. Errors in the initial pose estimation, for example, due to strong motion blur cause challenges in pose correction. The underlying assumption that the non-rigid motion is pose-dependent often fails in scenarios with complex clothing and accessories, hair styles, and large limb movements. Our proposed \textit{pose-independent temporal deformation} helps to supplement the missing information in its pose-dependent counterpart.

To this end, we introduce FlexNeRF, a novel approach for jointly learning a \textit{pose-dependent motion field} and \textit{pose-independent temporal deformation} within the NeRF framework for modeling human motions.
Moreover, we introduce a novel cycle consistency loss in our framework, further capitalizing on the fact that our canonical pose corresponds to one of the captured frames. The consistency regularizes the estimated deformation fields by mapping back and forth between each view and the canonical pose. Moreover, the information content of any frame in a motion sequence has a strong similarity to that of its neighbours. Hence, we propose to utilize this contextual information present in a consecutive set of frames to aid learning by imposing a temporal consistency loss. We additionally regularize the training by adding a supplementary loss based on the segmentation masks. Our approach allows photorealistic rendering of a moving human even when sparse views are available, by supplementing the pose-dependent motion field with additional information during learning: (i) pose-independent temporal deformation with ample pixel-wise correspondences beyond the (typically 24) pose point-correspondences, and (ii) consistency constraints/losses. In summary, our paper makes the following contributions: 


\begin{itemize}
    \item We propose a novel approach to learn  pose-independent temporal deformation to complement the pose-dependent motion for modeling humans in video, using one of the views as the canonical view.
    
    \item We propose a novel cyclic-consistency loss to regularize the learned deformations.
    
    \item We propose a temporal-consistency loss to aid learning with contextual information present in neighbouring frames, as well as to maintain consistency across consecutive rendered frames.
    
    \item Our approach outperforms the state-of-the-art approaches, with significant improvement in case of sparse views.
    
\end{itemize}
    
    
    
    

\section{Related Work}
\label{sec:related_work}
\subsection{Neural Radiance Fields (NeRFs)}
NeRFs attempt to learn a scene representation for novel-view synthesis by modeling the radiance field with learnable functions. A variety of approaches have been proposed recently for neural rendering, exploiting voxel grids~\cite{Sitzmann2019DeepVoxelsLP, Dellaert2021NeuralVR}, neural textures~\cite{Shysheya2019TexturedNA, Thies2019DeferredNR}, point-clouds~\cite{Meshry2019NeuralRI, Aliev2020NeuralPG}, and neural implicit functions~\cite{Chen2019LearningIF, Park2019DeepSDFLC}.

The landscape of neural rendering changed with NeRF \cite{Mildenhall2020NeRFRS}, which proposed a simple, yet revolutionary approach for photorealistic novel-view synthesis of static scenes. NeRF attempts to map from the 5-d light fields to 4-d space consisting of color $c$ (RGB) and density $\sigma$: likelihood that the light ray at this 5-d co-ordinate is terminated by occlusion. Since the introduction of original NeRF formulation, several variations and improvements \cite{Srinivasan2021NeRVNR, Zhang2021NeRFactorNF, Rebain2021DeRFDR, Liu2020NeuralSV} have been proposed.

\vspace{-0.5mm}
\subsection{Neural Rendering of Dynamic Scenes}

While originally proposed for static scenes, NeRF based approaches have been recently extended to dynamic scenes, for both rigid and non-rigid objects. These approaches can be divided into two main categories: a) optimizing a canonical configuration, and b) directly optimizing the 4-D spatio-temporal scenes. D-NeRF~\cite{Pumarola2021DNeRFNR} is an example of the first category, which attempts to map each observed frame to a given canonical frame. Once the canonical scene has been optimized for all available views, the novel-view can be rendered from the canonical space, and mapped back to the observed space. The same approach can be seen applied to videos with simple motion and other settings \cite{Park2020DeformableNR, Chen2021AnimatableNR, Tretschk2021NonRigidNR}. In contrast, the approaches that directly estimate spatio-temporal scene representations \cite{Xian2021SpacetimeNI, Li2021NeuralSF} takes positional-encoded or latent-coded time $t$ as an input in-addition to the spatial inputs, and attempts to predict the color and the density along each ray.

\vspace{-0.5mm}
\subsection{Neural Rendering of Human Subjects}
Compared to general rendering of dynamic scenes, human subject-specific rendering has additional challenges in terms of complex non-rigid human motions. Priors such as human pose that can provide additional information for successful scene representation. Hence, most methods \cite{Wu2020MultiViewNH, Peng2021AnimatableNR, Peng2021NeuralBI} begin with assuming \textit{SMPL} template as a prior \cite{Loper2015SMPLAS}. Furthermore, most methods use multi-view videos \cite{Xu2021HNeRFNR, Noguchi2021NeuralAR, Liu2021NeuralAN, Weng2020Vid2ActorFA}. A few recent methods including HumanNeRF \cite{Weng2022HumanNeRFFR} and others \cite{Gao2021DynamicVS, Chen2021AnimatableNR, Tretschk2021NonRigidNR} use monocular videos, whereas only the former attempts free-viewpoint rendering. However, these approaches have challenges rendering photorealistic outputs with sparse input views. We consider HumanNeRF\cite{Weng2022HumanNeRFFR} as the closet work to ours and address the aforementioned challenges.
\vspace{-1mm}
\section{Method}
\label{sec:method}
\begin{figure*}[t]
  \centering
   \includegraphics[width=0.8\linewidth]{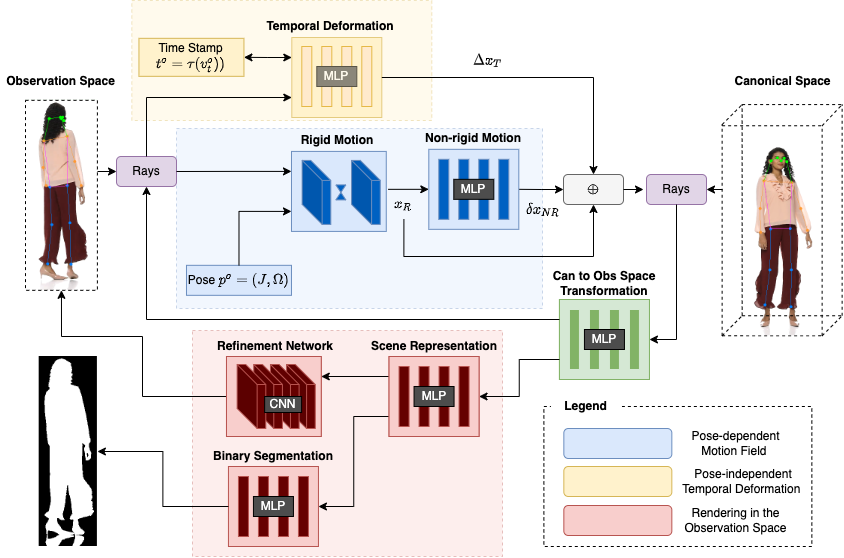}
   \caption{Overview of our approach. Pose-independent temporal deformation is used in conjunction with pose-dependent motion fields (rigid and non-rigid). We choose  one of the input frames as the canonical view, allowing us to use cyclic-consistency for regularization. }
   \label{fig:overview}
   \vspace{-1.25em}
\end{figure*}

Given a sequence of frames of a monocular video with a human manifesting complex motions, our goal is to achieve photo-realistic free-viewpoint rendering and reposing. We choose a frame as the canonical configuration (\eg the mid-point of the motion sequence) and learn it via: a) pose-dependent (rigid and non-rigid) motion fields and b) pose-independent temporal deformations. 

\subsection{Pose-Dependent Motion Fields}

Given a canonical pose-configuration $p^c = (J^c,\Omega^c)$ and the observed pose $p = (J,\Omega)$, where $\Omega$ represents the local joint rotations and $J$ represents the joint locations in 3D, we define a pose-guided motion field mapping between the observed and canonical spaces. We first compute the transformation $M_k(p^c, p)$, and hence the translation $t_k$ and rotation $R_k$ matrices between the joint coordinates in observed and canonical spaces, for a given body part $k$. $Y(w_i, j_i)$ computes the exponent of the local joint rotation $w_i$ of the joint location $j_i$ using the Rodrigues's formula \cite{Sorgi2011TwoViewGE},
\begin{align}
    Y(\omega_i, j_i) = \prod_{i \in \tau(k)} \begin{bmatrix} exp(\omega_i) & j_i \\ 0 & 1 \end{bmatrix},
\end{align}
where $\tau(k)$ denotes the ordered set of parents of the $k^{\text{th}}$ local joint. Subsequently, we compute the corresponding translation $t_k$ and rotation $R_k$ matrices,
\begin{align}
    \vspace{-3mm}
    M_k(p^c, p) = Y(\omega_i^c, j_i^c) \left\{ Y(\omega_i, j_i)  \right\}^{-1} = \begin{bmatrix} R_k & t_k \\ 0 & 1 \end{bmatrix}.
    \vspace{-3mm}
\end{align}

Given the translation and rotation matrices, we compute the rigid deformation $x_R$ between the observed and canonical spaces by defining
\begin{equation}
\vspace{-0.25em}
    \label{eq: l_x}
    \mathcal{L}(x) = \sum^K_{k=1} w^c_k (R_kx+t_k),
    \vspace{-0.25em}
\end{equation}
%
which represents the likelihood that the position $x$ is a part of the subject. We obtain the set of blend weight volumes in the canonical space $\{w_c^k\}^K_{k=1}$, where $K$ is the total number of 3D joint locations. To this end, starting from a constant random latent vector $z$, we generate the motion weight volume $W^c(x) = CNN_{\theta_{R}}(x;z) \in \mathbb{R}^4$ by optimizing the parameters $\theta_R$ of the $CNN_{\theta_R}$~\cite{Weng2022HumanNeRFFR}. We add a computed approximate Gaussian bone volume as a motion weight volume prior to the output of the last transposed convolution layer before activation. Subsequently, we compute the rigid deformation $x_R$ with the obtained $\mathcal{L}(x)$ and $W^c(x)$,
\begin{equation}
    \vspace{-1mm}
    \label{eq: x_r}
    x_{R}=\frac{\sum_{k=1}^{K}w^c_k(R_kx^o+t_k)^2}{\mathcal{L}(x)}.
\end{equation}

The non-rigid deformation between the observed and canonical spaces is then computed as a pose-guided offset $\delta x_{NR}$ to the rigid deformation $x_R$. We feed the positional encoding $\tau(x_R)$ to the non-rigid motion MLP as,
\begin{equation}
\vspace{-1mm}
    \delta x_{NR} = MLP_{\theta_{NR}}(\gamma(x_{R});\Omega).
\end{equation}

We follow the approach defined in \cite{Mildenhall2020NeRFRS} to obtain the positional encoding $\tau(x)$ of the position $x$. The non-rigid motion MLP consists of six fully-connected layers with the positional encoding $\tau(x_R)$ and the local joint rotations $\Omega$ (without global rotation) as the inputs with $\tau(x_R)$ skip-connected to the fifth layer to generate the offset. Since the initial pose estimate $p$ obtained from off-the-shelf techniques such as SPIN \cite{Kolotouros2019LearningTR} or VIBE \cite{Kocabas2020VIBEVI} can be erroneous, we perform a pose correction following \cite{Weng2022HumanNeRFFR}. 
\vspace{-0.25em}
\subsection{Pose-Independent Temporal Deformation}

We strategically set the canonical configuration to an observed frame in the training set, allowing us to access observed ($x^o$) and canonical ($x^c$) positions as a source of information. Furthermore, having a common canonical anchor when learning a dynamic setting ensures that the scene is inter-connected across frames and no longer independent between time instances, which is intuitive and known to provide quality performance \cite{Pumarola2021DNeRFNR}. This grounding aids the model to learn preserving temporal consistency of the dynamic scene. Nevertheless, such an approach which optimizes a canonical time configuration does not work well alone for free-viewpoint rendering where we are required to render a $360^0$ camera path with complex motion. Hence, we utilize a combined approach of pose-guided and pose-independent (time-guided) canonical configuration optimization. Results in Sec.~\ref{sec:results} show that the combined approach allows high quality photorealistic rendering with sparse views. 


We compute the pose-independent temporal deformation between a point position in the observation space $x^o$ to the canonical space $x^c$ with a temporal deformation MLP, similar to D-NeRF \cite{Pumarola2021DNeRFNR}. This temporal deformation $\Delta x_T$ is defined by,
\begin{equation}
    \vspace{-1mm}
    \Delta x_{\scriptscriptstyle T} = MLP_{\theta_{TD}}(\gamma(x^{o}),\gamma(x^{c}); (t^o, t^c)),
\end{equation}
where $t^o$ is the observed time stamp defined by $(t^o = \tau(v_t^o))$, and $t^c$ is the canonical time stamp defined by $(t^c = \tau(v_t^c))$. $v_t \in R^5$ is a learnable vector representation initialised proportional to the frame sequence index of the monocular video. 

In contrast to D-NeRF \cite{Pumarola2021DNeRFNR}, we set the temporal vectors to be learnable due to several reasons. Even though the progression of frame sequence indices are linear, the progression of temporal information throughout a video is highly non-linear. For instance, there can be rapid motion between two consecutive frames in a video, whereas there can be no motion between another two consecutive frames in the same video. Hence, it is not intuitive to allocate a linear representation to the temporal vectors $\{v_t\}$. Furthermore, albeit being a contrasting approach to ours, DyNeRF \cite{Li2022Neural3V} presents strong evidence that trainable (latent) codes can better handle complex scene dynamics such as large deformations and topological changes. We heavily regularize the training of these temporal vectors in order to ensure that they are contained within practical limits.

The temporal deformation MLP, $MLP_{\theta_{TD}}$ consists of 8 fully connected layers with the positional encoded temporal vectors $\tau(v_t^o))$ and $\tau(v_t^c))$, and the positional encoded point position vectors $\tau(x^o)$ and $\tau(x^c)$ as inputs. The observed encodings $\tau(v_t^o))$ and $\tau(x^o)$ are skip connected to the fifth layer to generate the deformation $\Delta x_T$. Finally, we aggregate the pose-guided rigid motion $x_R$, pose-guided non-rigid motion $\delta x_{NR}$ (as an offset to $x_R$), and the pose-independent temporal deformation $\Delta x_T$ to produce the predicted canonical configuration $\hat{x}^c$,
\begin{equation}
    \hat{x}^c = \underbrace{(x_R+\delta x_{NR})}_{\substack{\text{pose-guided motion}\\\text{field}}}  \ + \underbrace{\Delta x_T}_{\substack{\text{pose-independent}\\\text{temporal deformation}}}
\end{equation}

\subsection{Cyclic Consistency}

Having obtained the predicted canonical configuration $\hat{x}^c$, we predict the RGB color $c$ and the density $\sigma$ at a given spatial location. Rather than directly predicting ($c, \sigma$) from the canonical space similar to the existing approaches \cite{Weng2022HumanNeRFFR, Pumarola2021DNeRFNR, Xu2021HNeRFNR}, we propose to break the prediction process in two steps: a) transformation from canonical ($\hat{x}^c$) to observed ($\hat{x}^o$) space, and b) $(c, \sigma)$ prediction from $\hat{x}^o$. 

The proposed approach yields the opportunity to enforce a cyclic consistency constraint (observed $x^o$ $\rightarrow$ canonical $\hat{x}^c$ $\rightarrow$ observed $\hat{x}^o$) on the output of the canonical to observed transformation MLP, $\hat{x}_o = MLP_{\theta_{CO}}(\gamma(\hat{x}^{c}))$. Furthermore, having two separate specialized networks rather than one network to map from the rays in the canonical space to $(c,\sigma)$ in the observation space is more flexible and is empirically more effective as shown in Sec. \ref{sec:results}. 

The $MLP_{\theta_{CO}}$ has a similar architecture to $MLP_{\theta_{TD}}$, without the temporal vectors as inputs. The subsequent scene representation MLP, $(c,\sigma) = MLP_{\theta_{c}}(\gamma(\hat{x}^{o}))$ has a similar architecture to the network proposed in \cite{Mildenhall2020NeRFRS}. Prior to feeding $\hat{x}^c$ and $\hat{x}^o$ to the corresponding networks, each vector is positional encoded.

\begin{table*}[t]
\begin{center}
\begin{tabular}{c|c:c}
\multirow{2}{1em}{Input} & Novel View 1 & Novel View 2\\
\cline{2-3}
& HumanNeRF\cite{Weng2022HumanNeRFFR} \hspace{8mm} Ours \hspace{10mm} Ground Truth & HumanNeRF\cite{Weng2022HumanNeRFFR} \hspace{8mm} Ours \hspace{10mm} Ground Truth  \\
\hline \\
\includegraphics[width=0.14\linewidth, trim={0 0cm 0 0cm},clip] {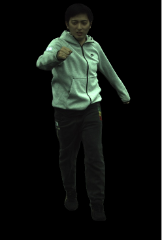} & \includegraphics[width=0.4\linewidth, trim={0 0.4cm 0 0cm},clip]{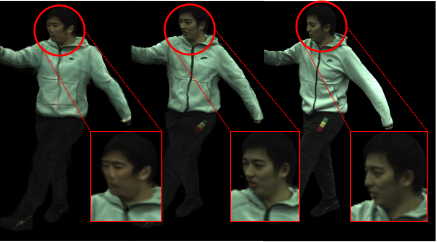} & \includegraphics[width=0.4\linewidth, trim={0 0.4cm 0 0cm},clip]{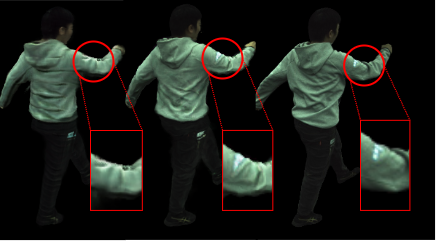} \\  
\hline
\includegraphics[width=0.14\linewidth, trim={0 0cm 0 0cm},clip] {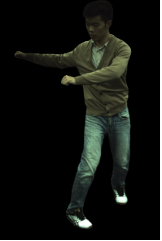} & \includegraphics[width=0.4\linewidth, trim={0 0.4cm 0 0cm},clip]{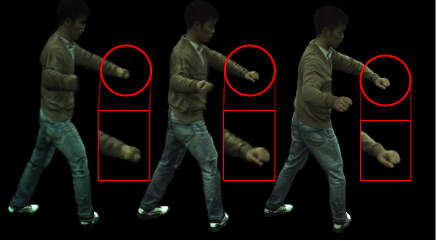} & \includegraphics[width=0.4\linewidth, trim={0 0.4cm 0 0cm},clip]{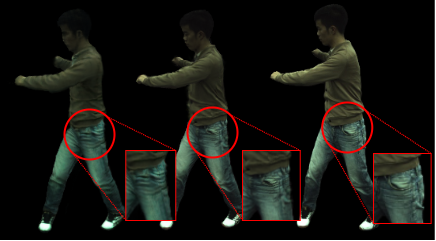} \\   
\hline
\includegraphics[width=0.14\linewidth, trim={0 0cm 0 0cm},clip] {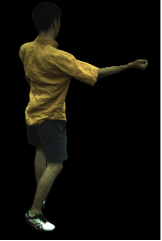} & \includegraphics[width=0.4\linewidth, trim={0 0.4cm 0 0cm},clip]{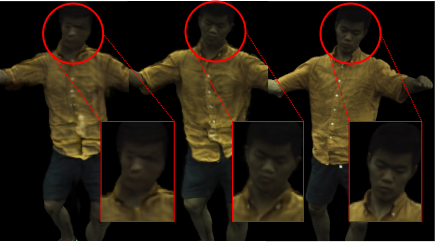} & \includegraphics[width=0.4\linewidth, trim={0 0.4cm 0 0cm},clip]{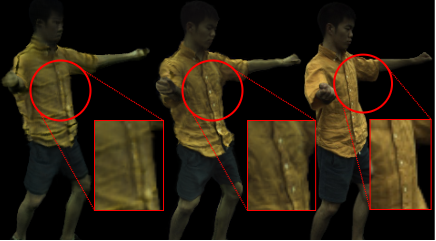} \\  
\hline
\end{tabular}
\end{center}
  \captionof{figure}{Qualitative comparison of rendered novel views on the ZJU-MoCap dataset. Notice the higher quality of rendered images from our method on details such as faces, buttons on shirt, etc.}
  \label{fig:mocap}
  \vspace*{-2ex}
\end{table*}

\subsection{Volume Rendering and Refinement Network}

We follow the volume rendering approach described in NeRF \cite{Mildenhall2020NeRFRS} by defining the expected alpha (density) mask $\mathcal{A}(r)$ and the expected color $\mathbf{C}(r)$ for a give ray $r$,
\begin{align}
    \mathcal{A}(r) &= \sum^{D}_{i=1} \left\{\prod^{i-1}_{j=1}(1-\alpha_j)\right\}\alpha_i \\
    C(r) &= \sum^{D}_{i=1} \left\{\prod^{i-1}_{j=1}(1-\alpha_j)\right\}\alpha_i c(x_i) \\
    \alpha_i &= \mathcal{L}(x_i)\{1-\text{exp}(-\sigma(x_i)\Delta z_i)\},
\end{align}
where $D$ is the number of samples, and $\Delta z_i$ is the interval between consecutive samples. We employ the same stratified sampling approach described in \cite{Mildenhall2020NeRFRS}. 

To further enhance the photorealism of the rendered images, we use a refinement network $\hat{I}_o = CNN_{\theta_{FT}}(C(r),\mathcal{A}(r))$ to add fine-grained details to the rendered image, similar to latent diffusion approaches \cite{Rombach2022HighResolutionIS}. The refinement network $CNN_{\theta_{FT}}$ consists three transposed convolution layers and outputs the final rendered image $\hat{I}_o$.




\subsection{Rendering Segmentation Mask}
The segmentation masks for the input frames can be obtained using an off-the-shelf segmentation network~\cite{He2020GrapyMLGP}. We use them to apply an additional loss to improve the density estimation. Note that rendering $\mathcal{A}(r)$ results in the predicted segmentation $\hat{M} =\mathcal{A}(r)$, which is compared against the real segmentation mask $M$. This helps to eliminate the halo effects~\cite{Weng2022HumanNeRFFR, Peng2021NeuralBI} and provide sharper boundaries. Empirically, we observed that thresholding the predicted segmentation mask, $\hat{M} =\mathcal{A}(r)H(\mathcal{A}(r),b)$ works better, where $b$ is a threshold value and \begin{equation} 
    H(\mathcal{A}(r),b)=\begin{cases} 1 \ \text{if} \ \mathcal{A}(r) > b \\ 0 \ \text{otherwise.} \end{cases}
    \label{eq: thresh_raw}
    \vspace{-0.25em}
\end{equation}

However, using a fixed threshold $b$ makes learning difficult at the start of training. To ease learning, we make $b$ a learnable parameter and re-define
$\hat{M} =(\mathcal{A}(r)+b)H(\mathcal{A}(r),b)$, so that it is differentiable with respect to $b$. We observe that $b$ goes to $0$ as training progresses, as in the ideal case. Compared to previous approaches such as \cite{Jiang2022NeuManNH}, this does not require us to depend on estimated depths, which could themselves be erroneous due to complex non-rigid motions.

\vspace{-2mm}
\section{Learning the FlexNeRF model}
\label{sec:learning}

\begin{table*}[ht!]
\begin{center}
\begin{tabular}{ c| c : c: c}
\multirow{2}{1em}{Input} & Novel View 1 & Novel View 2 & Novel View 3\\
\cline{2-4}
& \hspace{-4mm} HumanNeRF \hspace{1mm} Ours &  \hspace{-4mm} HumanNeRF \hspace{1mm} Ours & \hspace{-4mm} HumanNeRF \hspace{1mm} Ours   \\
\hline \\
\includegraphics[width=0.07\linewidth, trim={0 0cm 0 0cm},clip] {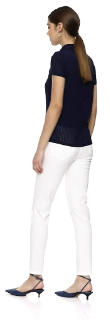} & \includegraphics[width=0.2\linewidth, trim={0 2cm 0 0.2cm},clip]{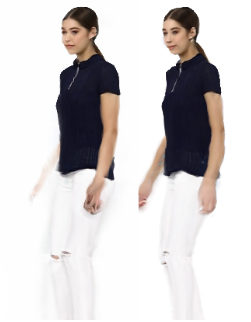} & \includegraphics[width=0.2\linewidth, trim={0 2cm 0 0.2cm},clip]{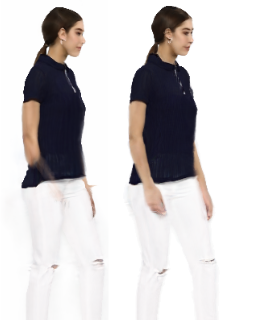} & 
\includegraphics[width=0.2\linewidth, trim={0 2cm 0 0.2cm},clip]{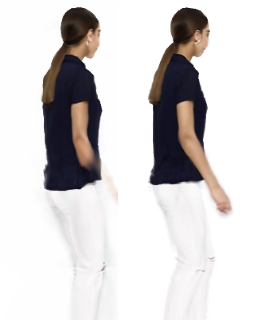}\\  
\hline
\includegraphics[width=0.07\linewidth, trim={0 0cm 0 0cm},clip] {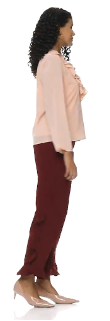} & \includegraphics[width=0.2\linewidth, trim={0 2.2cm 0 0},clip]{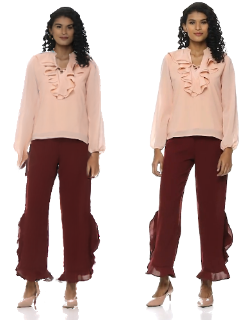} & \includegraphics[width=0.2\linewidth, trim={0 2cm 0 0.2cm},clip]{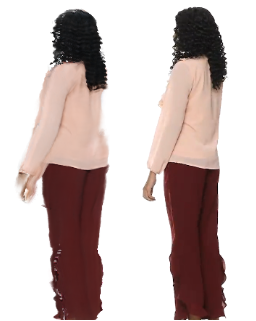} & 
\includegraphics[width=0.2\linewidth, trim={0 2cm 0 0.2cm},clip]{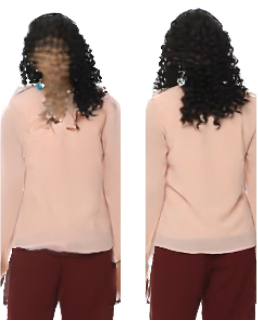}\\  
\hline
\includegraphics[width=0.069\linewidth, trim={0 0cm 0 0cm},clip] {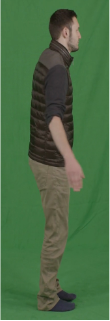} & \includegraphics[width=0.194\linewidth, trim={0 2cm 0 0.2cm},clip]{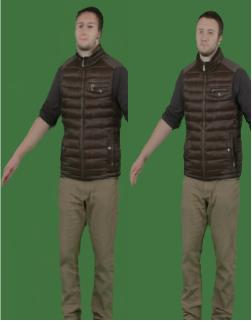} & \includegraphics[width=0.199\linewidth, trim={0 2cm 0 0.2cm},clip]{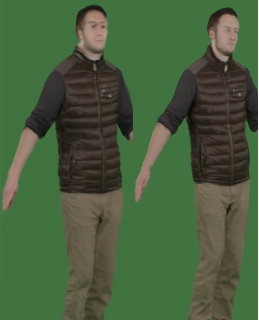} & 
\includegraphics[width=0.197\linewidth, trim={0 2cm 0 0.2cm},clip]{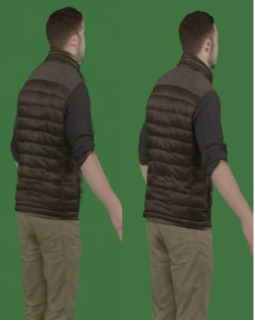}\\  
\hline

\end{tabular}
\end{center}
  \captionof{figure}{Qualitative comparison of novel rendered views on SCF dataset (top two rows) and the People Snapshot dataset (bottom row) using sparse views. Our approach significantly improves the results.}
  \label{fig:itw}
  \vspace*{-1ex}
\end{table*}

In this section we describe the loss functions used to learn the FlexNeRF model and discuss details with respect to optimization and ray-sampling.
\vspace{-1mm}
\subsection{Loss Functions}
NeRFs are typically trained with a combination of losses between the rendered and observed frames. In addition, FlexNeRF also uses a combination of segmentation loss, cyclic consistency loss and temporal consistency loss as defined below.

\noindent\textbf{Segmentation Loss:} We apply the BCE-Dice loss between the predicted and ground truth binary segmentation masks
\begin{equation}
\begin{aligned}
    \mathbb{L}_{S} &= \frac{1}{N} \sum \left[ M \text{log} \hat{M} + (1-M)\text{log}(1-\hat{M}) \right] \\
    &+ \frac{2|M \cap \hat{M}|}{|M|+|\hat{M}|},
\end{aligned}
\end{equation}
where $N$ is the number of pixels in the segmentation mask. 

\begin{figure}[t]
    \vspace*{-3ex}
  \centering
   \includegraphics[width=0.85\linewidth]{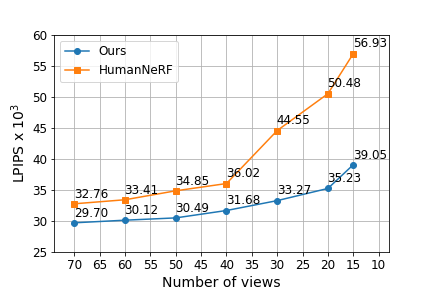}
   \caption{LPIPS metric comparison on ZJU-MoCap between HumanNeRF \cite{Weng2022HumanNeRFFR} and our method with decreasing number of views.}
   \label{fig:views_comp}
   \vspace*{-4ex}
\end{figure}

\noindent\textbf{Cyclic Consistency Loss:} We introduce a cyclic consistency constraint on the canonical to observation space transformation, using Mean Squared Error (MSE) between $\hat{x}^o$ and $x^o$ defined by,
\begin{equation}
    \mathbb{L}_{CCL} = \frac{1}{L} \sum_{i=1}^{L} (\hat{x}^o_i- x^o_i)^2,
\end{equation}
where $L$ is the number of positional samples. 
\vspace{1mm}

\begin{table*}[t]
\begin{center}
\begin{tabular}{ |c|c|c|c|c|c| } 
\hline
{Dataset} & {Views} & {Method} & {LPIPS} $\times 10^3$ $\downarrow$ & {PSNR} $\uparrow$ & {SSIM} $\uparrow$\\ 
\hline
\rowcolor{gray!25}
\cellcolor{white} \multirow{6}{8em}{PeopleSnapshot \cite{alldieck2018video}} &  & HumanNeRF \cite{Weng2022HumanNeRFFR} & 39.27 & 27.65 & 0.8816 \\ 
\hhline{~~|*4{-}|}
\rowcolor{gray!25}
\cellcolor{white} & \multirow{-2}{5em}{Sparse$^\ddagger$} & Ours & \textbf{37.11} & \textbf{28.09} & \textbf{0.9003} \\ 
\cline{2-6}
 & \multirow{3}{5em}{Full} & Neural Body \cite{Peng2021NeuralBI} & 57.67$^*$ & 24.62 & 0.8490 \\ 
\cline{3-6}
& & HumanNeRF \cite{Weng2022HumanNeRFFR} & 36.79 & 28.05 & 0.8984 \\ 
\cline{3-6}
& & Ours & \textbf{35.63} & \textbf{28.77} & \textbf{0.9043} \\
\hline
\hline
\rowcolor{gray!25}
\cellcolor{white} \multirow{6}{8em}{ZJU-MoCap \cite{Peng2021NeuralBI, fang2021mirrored}} &  & HumanNeRF \cite{Weng2022HumanNeRFFR} & 36.02 & 29.82 & 0.9597 \\ 
\arrayrulecolor{gray!25}
\hhline{~|*1{-}|}
\arrayrulecolor{black}
\hhline{~~|*4{-}|}
\rowcolor{gray!25}
\cellcolor{white} & \multirow{-2}{5em}{Sparse$^\ddagger$} & Ours & \textbf{31.68} & \textbf{30.18} & \textbf{0.9685} \\ 
\cline{2-6}
 & \multirow{3}{5em}{Full} & Neural Body \cite{Peng2021NeuralBI} & 52.28 & 29.07 & 0.9615 \\ 
\cline{3-6}
\cline{3-6}
& & HumanNeRF \cite{Weng2022HumanNeRFFR} & 31.72 & 30.24 & 0.9679 \\ 
\cline{3-6}
& & Ours & \textbf{29.01} & \textbf{31.73} & \textbf{0.9765} \\

\hline
\hline
\rowcolor{gray!25}
\cellcolor{white} &  & Neural Body \cite{Peng2021NeuralBI} & 48.62 & 25.07 & 0.9131 \\ 
\cline{3-6}
\rowcolor{gray!25}
\hhline{~~|*4{-}|}
\rowcolor{gray!25}
\cellcolor{white} & & HumanNeRF \cite{Weng2022HumanNeRFFR} & 39.71 & 26.12 & 0.9366 \\ 
\hhline{~~|*4{-}|}
\rowcolor{gray!25}
\cellcolor{white} \multirow{-4}{8em}{SCF Dataset$^\dag$} & \multirow{-3}{5em}{\cellcolor{gray!25} Sparse$^\ddagger$} & Ours & \textbf{34.26} & \textbf{29.55} & \textbf{0.9627} \\
\hline
\end{tabular}

\end{center}
  \caption{Comparison of performance across benchmark datasets. $*$ refers to adjusted LPIPS from the values reported in~\cite{Xu2021HNeRFNR} to fit the same scale as our experiments. $\dag$ refers to the Self-Captured Fashion (SCF) dataset. $\ddagger$ indicates the model trained with sparse ($\sim 40$) views.}
  \label{tab:all_datasets}
\end{table*}

\noindent\textbf{Temporal Consistency Loss (TCL):} We identify that imposing temporal consistency constraints can be valuable at two instances: a) while rendering consecutive training frames $\{\hat{I}^o\}_{t=-k}^{k}$ and b) while applying temporal deformation from consecutive training frames to the canonical frame $\{{\Delta x}_T\}_{t=-k}^{k}$. To this end, we employ the cycle-back regression consistency loss proposed in \cite{Dwibedi2019TemporalCL}. The cycle-back regression attempts to determine the temporal proximity of rendered frames or deformation vectors, and penalize the model if they are not in close temporal proximity. Given a rendered frame or a deformation vector $u$, and neighbors $\{v_k\}$, we compute the similarity vector $\beta_k$,
\begin{equation}
    \vspace{-1ex}
    \beta_k = \frac{\text{exp}(-\|u-v_k\|^2)}{\sum_j \text{exp}(-\|u-v_j\|^2)},
\end{equation}
where $u, v_k \in \{v_k\}$ and $\beta$ is a discrete distribution of similarities over time. We impose a Gaussian prior on $\beta$ by minimizing the normalized square distance, 
\begin{align}
    \mu = \sum_k k \beta_k \ \ \ \ \sigma^2 = \beta_k (k-\mu)^2
\end{align}
\vspace{-9mm}
\begin{align}
    \mathbb{L}_{TCL} = \frac{|i-\mu|^2}{\sigma^2} + \lambda log(\sigma),
\end{align}

where $\lambda$ is a regularization parameter. 
Finally, the rendering loss
$\mathbb{L}_{rend}$, the canonical loss  $\mathbb{L}_{can}$, and the overall loss $\mathbb{L}$ are defined as 
\vspace{-3mm}
\begin{align}
\mathbb{L}_{rend} &= \mathbb{L}_{LPIPS}(\hat{I}^o,I^o) + \mathbb{L}_{MSE}(\hat{I}^o,I^o) \\
\nonumber &+ \mathbb{L}_{TCL}(\{\hat{I}^o\}_{t=-k}^{k})\\
\mathbb{L}_{can} &= \mathbb{L}_{MSE}(\hat{x}^c,x^c) + \mathbb{L}_{TCL}(\{{\Delta x}_T\}_{t=-k}^{k})\\
\mathbb{L} &= \mathbb{L}_{rend} + \mathbb{L}_{can} + \mathbb{L}_{CCL} + \mathbb{L}_{S}
\end{align}


\subsection{Optimization Details}
\vspace{-1mm}
\noindent\textbf{Delayed Modular Optimization:} We follow a delayed-optimization approach similar to \cite{Weng2022HumanNeRFFR} to optimize the non-rigid motion, binary segmentation, and the refinement modules of our method. Optimizing these modules from the beginning yields lower performance as they rely on adequate inputs from the rest of the system. Hence, we freeze these modules initially, and unfreeze them gradually during the course of training.

\noindent\textbf{Ray Sampling:} Since LPIPS use a convolution-based approach to extract features, we use patch-based ray sampling following \cite{Weng2022HumanNeRFFR, Schwarz2020GRAFGR} instead of random ray sampling~\cite{Mildenhall2020NeRFRS} from the whole image. 

\vspace{-2mm}
\section{Experiments and Results}
\label{sec:results}
\vspace{-0.5mm}
\subsection{Benchmark Datasets and Metrics}
\vspace{-0.5mm}
We evaluate the proposed method on two public datasets: ZJU-MoCap \cite{Peng2021NeuralBI, fang2021mirrored} and People Snapshot \cite{alldieck2018video}, and one Self-Captured Fashion (SCF) dataset. The People Snapshot dataset has 7 sequences of monocular videos of human subjects displaying rotating motions in front of a static camera. For the ZJU-MoCap dataset, we use 6 sequences to be compatible with \cite{Weng2022HumanNeRFFR} for comparison purposes. We use only the views from the first camera to simulate monocular video settings for training, and use the views from rest of the cameras for evaluation. Videos from both ZJU-MoCap and People Snapshot datasets are carefully captured under lab settings. 


For the SCF dataset, we captured 7 sequences of monocular videos freely where the movements are solely up to the discretion of the consenting subjects. In contrast to the public datasets: a) the subjects are wearing complex clothing and accessories, and the movements are fast, b) videos are captured without any controlled settings, and c) the captured videos are brief with only one full rotation. Due to the absence of ground truths for the SCF and People Snapshot datasets, we evaluate the rendered frames by holding random frames from training. We use LPIPS, PSNR, and SSIM \cite{Zhang2021NeRFactorNF} as evaluation metrics.

\begin{table}[t]
\begin{center}
\begin{tabular}{c}
\hspace{-6mm} HumanNeRF \hspace{1mm} Ours \hspace{2mm} HumanNeRF \hspace{6mm} Ours \hspace{8mm}  \\
\hline \\
\includegraphics[width=0.75\linewidth, trim={0.4cm 0 1.4cm 0},clip]{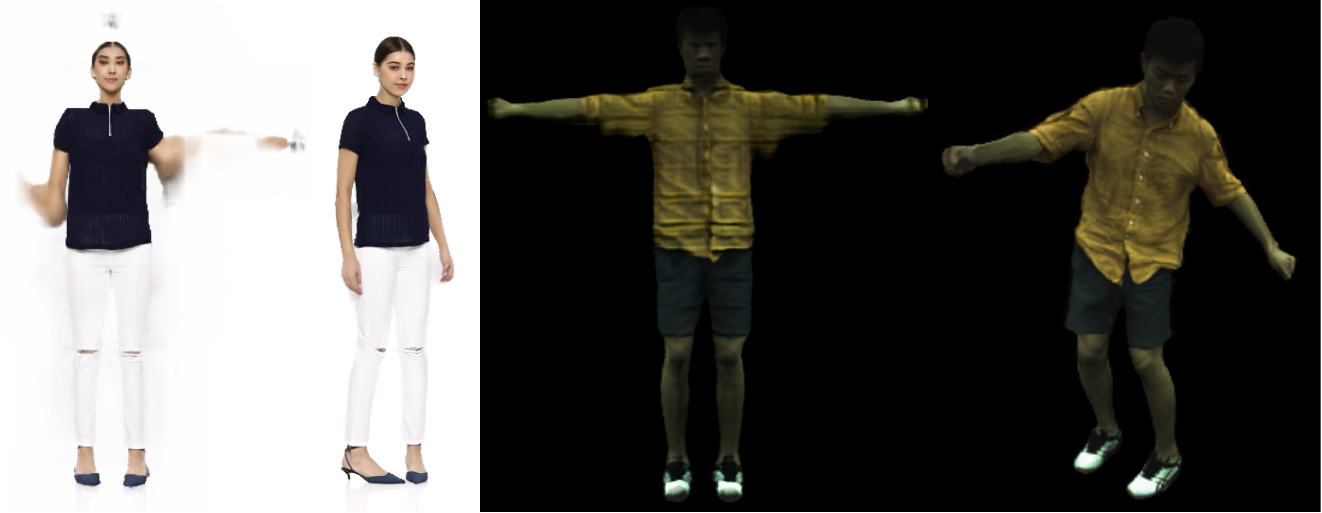} 
\vspace*{-5ex}
\end{tabular}

\end{center}
  \captionof{figure}{Comparing the rendering of the canonical view for SCF (left) and ZJU-Mocap (right) datasets. Our approach is able to learn a higher quality canonical view.}
  \label{fig:analysis}
  \vspace*{-3ex}
\end{table}
\vspace{-0.5mm}
\subsection{Results and Analysis}
\vspace{-0.5mm}



\definecolor{amber}{rgb}{1.0, 0.75, 0.0}
\begin{table*}[t!]
\small
\begin{center}
\begin{tabular}{ |c|c:c|c:c|c:c| } 
\hline
\textbf{Ablation} & \textbf{{LPIPS} $\times \mathbf{10^3}$ $\downarrow$} & $\Delta$ & \textbf{{PSNR} $\uparrow$} & $\Delta$ & \textbf{{SSIM} $\uparrow$} & $\Delta \times \mathbf{10^{-2}}$\\ 
\hline
HumanNeRF & 39.71 & 0.00 & 26.12 & 0.00 & 0.9366 & 0.00\\ 
\hline
Ours (full) & \textbf{34.26} & \textbf{-5.45} & \textbf{29.55} & \textbf{+3.43} & \textbf{0.9627} & \textbf{+2.61}\\ 
\hline
Ours (w/o PID) & 37.12 & -2.59 & 27.42 & +1.30 & 0.9469 & +1.03\\
\hline
Ours (w/o CO-MLP + CCL) & 36.86 & -2.85 & 28.50 & +2.38 & 0.9509 & +1.43 \\
\hline
Ours (w/o TCL) & 36.17 & -3.54 & 27.94 & +1.82 & 0.9521 & +1.55\\
\hline
Ours (w/o RF) & 35.04 & -4.67 & 29.17 & +3.05 & 0.9593 & +2.27\\
\hline
Ours (w/o BS) & 35.78 & -3.93 & 29.07 & +2.95 & 0.9601 & +2.35\\
\hline
\end{tabular}

\end{center}
\vspace{-6mm}
  \caption{Ablation: Effect of removing various modules and losses from our full approach on the SCF dataset. \textit{PID:} Pose-Independent temporal Deformation module, \textit{CO-MLP:} Canonical-to-Observed transformation MLP, CCL: Cyclic Consistency Loss, \textit{RF:} Refinement MLP, \textit{BS:} Binary Segmentation Loss, \textit{TCL:} Temporal Consistency Loss.}
  \label{tab:ablation}
\end{table*}

\begin{table}[t]
\begin{center}
\begin{tabular}{c}
 \hspace{-6mm} Worst case \hspace{14mm} Best Case  \\
\hline \\
\includegraphics[width=0.7\linewidth, trim={0.6cm 0 0.4cm 0},clip]{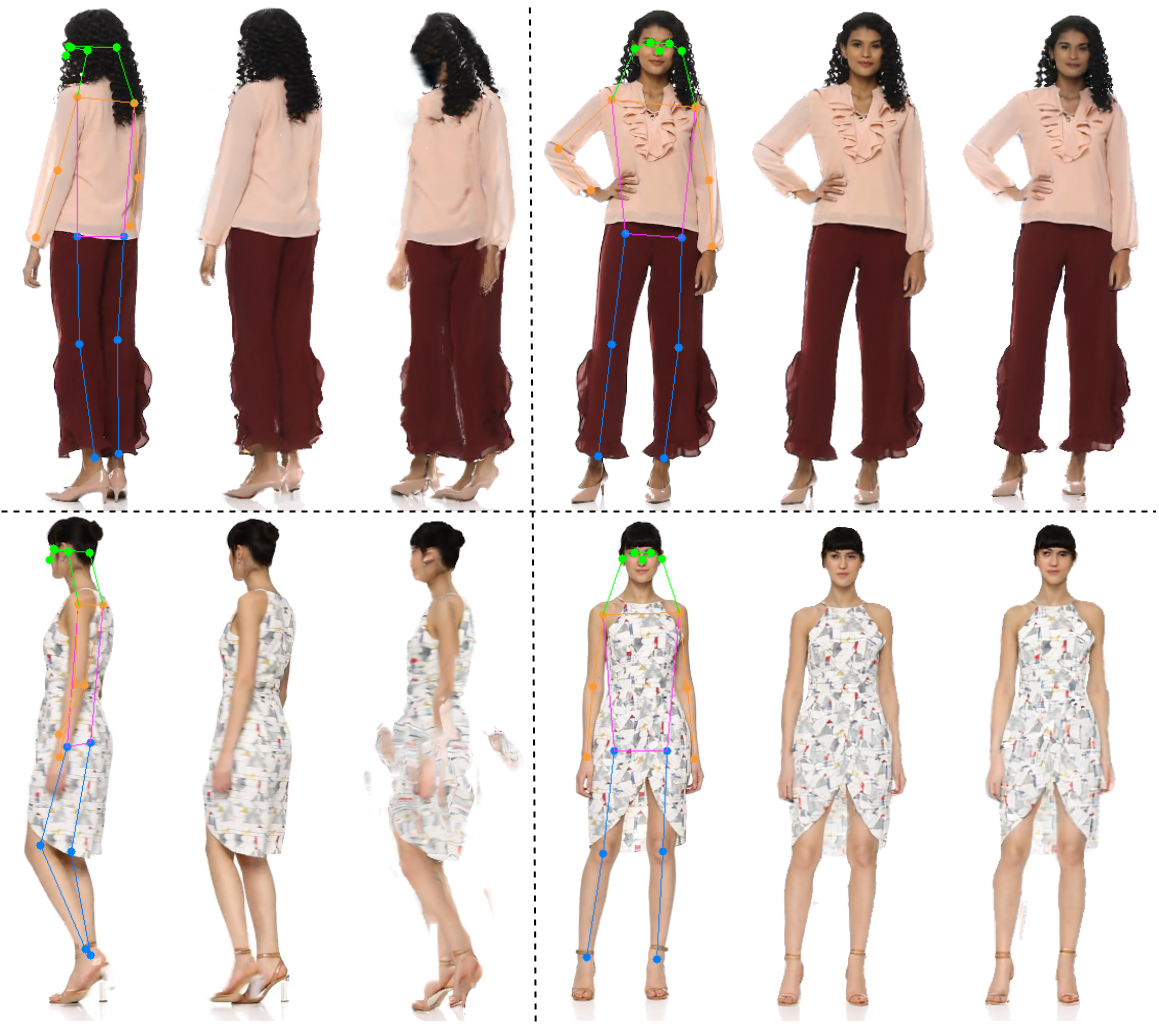}
\vspace*{-5ex}
\end{tabular}
\end{center}
  \captionof{figure}{Each cell shows the original training frame (left) with rendered frame using the \textit{same} viewpoint after training. Note that for difficult/challenging poses, HumanNeRF (right) fails to minimize the training loss compared to ours (middle). }
  \label{fig:render_training_frames}
  \vspace*{-4ex}
\end{table}

\vspace{-0.75mm}
\noindent\textbf{Quantitative Results:} Table~\ref{tab:all_datasets} compares our method against HumanNeRF~\cite{Weng2022HumanNeRFFR} and Neural-Body~\cite{Peng2021NeuralBI} across the three datasets. We consider two settings: \textit{Full}, using all the frames and \textit{Sparse}, using a sparse number of views. To generate the \textit{Sparse} setting, we remove stationary frames from the video and any subsequent frames after the first  complete rotation of the subject. Subsequently, we re-sample every $k^{\text{th}}$ frame, where $k$ is chosen such that $\sim8-10\%$ of the original number of frames remain. Our approach outperforms HumanNeRF and NeuralBody across all metrics in the \textit{Full} setting. Since HumanNeRF is significantly better than Neural Body using all the frames, we only compare with HumanNeRF for sparse view setting. Our approach also outperforms HumanNeRF in the \textit{Sparse} setting for all three datasets as shown in Table~\ref{tab:all_datasets}.



\noindent\textbf{Analysis of Number of Views:} To further analyze the effect of the number of views, we train various models using different number of input views. Figure~\ref{fig:views_comp} compares the performance of LPIPS metric for our method against HumanNeRF \cite{Weng2022HumanNeRFFR} with varying number of views. Our approach is better than HumanNeRF for all settings, with significant reduction in LPIPS metric as the number of views decreases.

\noindent\textbf{Quality of Canonical View:} An interesting analysis is to visualize the quality of the canonical view rendering itself. This indicates how well the model can learn deformations and fuse information from various frames. As shown in Figure~\ref{fig:analysis}, our approach can produce significantly higher quality of rendering for canonical view, thanks to our proposed pose-independent temporal deformation.

\noindent\textbf{Quality of Rendered Training View:} We can infer how well the NeRF model is minimizing the training loss by re-rendering the trained model using the same viewpoint as input training frames. Figure~\ref{fig:render_training_frames} shows the best and worst frames for training loss minimization. Notice that for easy poses (frontal), the rendered training frame from both HumanNeRF and our method are visually similar to the original training frame. However, for challenging poses (side view), the quality and accuracy of rendered training frame is significantly higher for our method compared to HumanNeRF. Infact, in the example shown in Figure~\ref{fig:render_training_frames} (top row), HumanNeRF totally fails to render the correct pose of the training frame itself, hindering overall learning. This again highlights the pitfalls of pose-dependent learning, and indicates that our method can better transform input frames to canonical frame, thanks to the proposed cyclic consistency constraint and pose-independent temporal deformation. Moreover, our method can render neighbouring frames that are not used for training in the \textit{sparse} setting by passing the interpolated frame index. However, both our method and HumanNeRF share similar challenges generating completely unseen poses \textit{beyond} the range in the input video.




\noindent\textbf{Qualitative Results:} Figs.~\ref{fig:mocap} and~\ref{fig:itw} show qualitative results for rendering from novel view-points for the ZJU-MoCap and SCF datasets respectively. Our approach renders higher quality novel-views compared to HumanNerf (on faces, buttons on t-shirts, etc.).

\noindent\textbf{Ablations}: Table~\ref{tab:ablation} shows the effect of removing various modules from our full approach. We observe that all the proposed losses and constraints contribute to the performance improvement.

\vspace{-2.5mm}
\section{Conclusion}
\label{sec:conclusion}
\vspace{-1.75mm}
In this work, we presented FlexNeRF: a novel method for photorealistic free-viewpoint rendering of moving humans starting with monocular videos. Our proposed framework utilizes pose-independent temporal deformation along with cycle consistency to better model the complex human motions with sparse input views. Experiments demonstrated that our method outperforms the state-of-the-art approaches in rendering novel viewpoints. We hope that our work will spark future research in the challenging problem of free-viewpoint rendering from in-the-wild videos.

\vspace{0.25mm}
\small
\noindent\textbf{Acknowledgments:} This project was partially funded by the DARPA SemaFor (HR001119S0085) program.


{\small
\bibliographystyle{ieee_fullname}
\bibliography{egbib}

\begin{thebibliography}{10}\itemsep=-1pt

\bibitem{Aliev2020NeuralPG}
Kara-Ali Aliev, Dmitry Ulyanov, and Victor~S. Lempitsky.
\newblock Neural point-based graphics.
\newblock {\em ArXiv}, abs/1906.08240, 2020.

\bibitem{alldieck2018video}
Thiemo Alldieck, Marcus Magnor, Weipeng Xu, Christian Theobalt, and Gerard
  Pons-Moll.
\newblock Video based reconstruction of 3d people models.
\newblock In {\em {IEEE}/{CVF} Conference on Computer Vision and Pattern
  Recognition ({CVPR})}, pages 8387--8397, Jun 2018.
\newblock {CVPR} Spotlight Paper.

\bibitem{Chen2021AnimatableNR}
Jianchuan Chen, Ying Zhang, Di Kang, Xuefei Zhe, Linchao Bao, and Huchuan Lu.
\newblock Animatable neural radiance fields from monocular rgb video.
\newblock {\em ArXiv}, abs/2106.13629, 2021.

\bibitem{Chen2019LearningIF}
Zhiqin Chen and Hao Zhang.
\newblock Learning implicit fields for generative shape modeling.
\newblock {\em 2019 IEEE/CVF Conference on Computer Vision and Pattern
  Recognition (CVPR)}, pages 5932--5941, 2019.

\bibitem{Dellaert2021NeuralVR}
Frank Dellaert and Yen-Chen Lin.
\newblock Neural volume rendering: Nerf and beyond.
\newblock {\em ArXiv}, abs/2101.05204, 2021.

\bibitem{Dwibedi2019TemporalCL}
Debidatta Dwibedi, Yusuf Aytar, Jonathan Tompson, Pierre Sermanet, and Andrew
  Zisserman.
\newblock Temporal cycle-consistency learning.
\newblock {\em 2019 IEEE/CVF Conference on Computer Vision and Pattern
  Recognition (CVPR)}, pages 1801--1810, 2019.

\bibitem{fang2021mirrored}
Qi Fang, Qing Shuai, Junting Dong, Hujun Bao, and Xiaowei Zhou.
\newblock Reconstructing 3d human pose by watching humans in the mirror.
\newblock In {\em CVPR}, 2021.

\bibitem{Gao2021DynamicVS}
Chen Gao, Ayush Saraf, Johannes Kopf, and Jia-Bin Huang.
\newblock Dynamic view synthesis from dynamic monocular video.
\newblock {\em 2021 IEEE/CVF International Conference on Computer Vision
  (ICCV)}, pages 5692--5701, 2021.

\bibitem{He2020GrapyMLGP}
Haoyu He, Jing Zhang, Qiming Zhang, and Dacheng Tao.
\newblock Grapy-ml: Graph pyramid mutual learning for cross-dataset human
  parsing.
\newblock In {\em AAAI}, 2020.

\bibitem{Jiang2022NeuManNH}
Wei Jiang, Kwang~Moo Yi, Golnoosh Samei, Oncel Tuzel, and Anurag Ranjan.
\newblock Neuman: Neural human radiance field from a single video.
\newblock {\em ArXiv}, abs/2203.12575, 2022.

\bibitem{Kocabas2020VIBEVI}
Muhammed Kocabas, Nikos Athanasiou, and Michael~J. Black.
\newblock Vibe: Video inference for human body pose and shape estimation.
\newblock {\em 2020 IEEE/CVF Conference on Computer Vision and Pattern
  Recognition (CVPR)}, pages 5252--5262, 2020.

\bibitem{Kolotouros2019LearningTR}
Nikos Kolotouros, Georgios Pavlakos, Michael~J. Black, and Kostas Daniilidis.
\newblock Learning to reconstruct 3d human pose and shape via model-fitting in
  the loop.
\newblock {\em 2019 IEEE/CVF International Conference on Computer Vision
  (ICCV)}, pages 2252--2261, 2019.

\bibitem{Li2022Neural3V}
Tianye Li, Miroslava Slavcheva, Michael Zollhoefer, Simon Green, Christoph
  Lassner, Changil Kim, Tanner Schmidt, S. Lovegrove, Michael Goesele,
  Richard~A. Newcombe, and Zhaoyang Lv.
\newblock Neural 3d video synthesis from multi-view video.
\newblock {\em 2022 IEEE/CVF Conference on Computer Vision and Pattern
  Recognition (CVPR)}, pages 5511--5521, 2022.

\bibitem{Li2021NeuralSF}
Zhengqi Li, Simon Niklaus, Noah Snavely, and Oliver Wang.
\newblock Neural scene flow fields for space-time view synthesis of dynamic
  scenes.
\newblock {\em 2021 IEEE/CVF Conference on Computer Vision and Pattern
  Recognition (CVPR)}, pages 6494--6504, 2021.

\bibitem{Liu2020NeuralSV}
Lingjie Liu, Jiatao Gu, Kyaw~Zaw Lin, Tat-Seng Chua, and Christian Theobalt.
\newblock Neural sparse voxel fields.
\newblock {\em ArXiv}, abs/2007.11571, 2020.

\bibitem{Liu2021NeuralAN}
Lingjie Liu, Marc Habermann, V. Rudnev, Kripasindhu Sarkar, Jiatao Gu, and
  Christian Theobalt.
\newblock Neural actor: Neural free-view synthesis of human actors with pose
  control.
\newblock {\em ArXiv}, abs/2106.02019, 2021.

\bibitem{Liu2020NeuralHV}
Lingjie Liu, Weipeng Xu, Marc Habermann, Michael Zollh{\"o}fer, Florian
  Bernard, Hyeongwoo Kim, Wenping Wang, and Christian Theobalt.
\newblock Neural human video rendering by learning dynamic textures and
  rendering-to-video translation.
\newblock {\em IEEE transactions on visualization and computer graphics}, PP,
  2020.

\bibitem{Loper2015SMPLAS}
Matthew Loper, Naureen Mahmood, Javier Romero, Gerard Pons-Moll, and Michael~J.
  Black.
\newblock Smpl: a skinned multi-person linear model.
\newblock {\em ACM Trans. Graph.}, 34:248:1--248:16, 2015.

\bibitem{MartinBrualla2018LookinGoodEP}
Ricardo Martin-Brualla, Rohit Pandey, Shuoran Yang, Pavel Pidlypenskyi,
  Jonathan Taylor, Julien P.~C. Valentin, S. Khamis, Philip~L. Davidson,
  Anastasia Tkach, Peter Lincoln, Adarsh Kowdle, Christoph Rhemann, Dan~B.
  Goldman, Cem Keskin, Steven~M. Seitz, Shahram Izadi, and S. Fanello.
\newblock Lookingood: Enhancing performance capture with real-time neural
  re-rendering.
\newblock {\em ACM Trans. Graph.}, 37:255, 2018.

\bibitem{MartinBrualla2021NeRFIT}
Ricardo Martin-Brualla, Noha Radwan, Mehdi S.~M. Sajjadi, Jonathan~T. Barron,
  Alexey Dosovitskiy, and Daniel Duckworth.
\newblock Nerf in the wild: Neural radiance fields for unconstrained photo
  collections.
\newblock {\em 2021 IEEE/CVF Conference on Computer Vision and Pattern
  Recognition (CVPR)}, pages 7206--7215, 2021.

\bibitem{Meshry2019NeuralRI}
Moustafa Meshry, Dan~B. Goldman, S. Khamis, Hugues Hoppe, Rohit Pandey, Noah
  Snavely, and Ricardo Martin-Brualla.
\newblock Neural rerendering in the wild.
\newblock {\em 2019 IEEE/CVF Conference on Computer Vision and Pattern
  Recognition (CVPR)}, pages 6871--6880, 2019.

\bibitem{Mildenhall2020NeRFRS}
Ben Mildenhall, Pratul~P. Srinivasan, Matthew Tancik, Jonathan~T. Barron, Ravi
  Ramamoorthi, and Ren Ng.
\newblock Nerf: Representing scenes as neural radiance fields for view
  synthesis.
\newblock In {\em ECCV}, 2020.

\bibitem{Noguchi2021NeuralAR}
Atsuhiro Noguchi, Xiao Sun, Stephen Lin, and Tatsuya Harada.
\newblock Neural articulated radiance field.
\newblock {\em 2021 IEEE/CVF International Conference on Computer Vision
  (ICCV)}, pages 5742--5752, 2021.

\bibitem{Ouyang2022RealTimeNC}
Hao Ouyang, Bo Zhang, Pan Zhang, Hao Yang, Jiaolong Yang, Dong Chen, Qifeng
  Chen, and Fang Wen.
\newblock Real-time neural character rendering with pose-guided multiplane
  images.
\newblock {\em ArXiv}, abs/2204.11820, 2022.

\bibitem{Park2019DeepSDFLC}
Jeong~Joon Park, Peter~R. Florence, Julian Straub, Richard~A. Newcombe, and S.
  Lovegrove.
\newblock Deepsdf: Learning continuous signed distance functions for shape
  representation.
\newblock {\em 2019 IEEE/CVF Conference on Computer Vision and Pattern
  Recognition (CVPR)}, pages 165--174, 2019.

\bibitem{Park2020DeformableNR}
Keunhong Park, U. Sinha, Jonathan~T. Barron, Sofien Bouaziz, Dan~B. Goldman,
  Steven~M. Seitz, and Ricardo~Mart{\'i}n Brualla.
\newblock Deformable neural radiance fields.
\newblock {\em ArXiv}, abs/2011.12948, 2020.

\bibitem{Peng2021AnimatableNR}
Sida Peng, Junting Dong, Qianqian Wang, Shang-Wei Zhang, Qing Shuai, Xiaowei
  Zhou, and Hujun Bao.
\newblock Animatable neural radiance fields for modeling dynamic human bodies.
\newblock {\em 2021 IEEE/CVF International Conference on Computer Vision
  (ICCV)}, pages 14294--14303, 2021.

\bibitem{Peng2021NeuralBI}
Sida Peng, Yuanqing Zhang, Yinghao Xu, Qianqian Wang, Qing Shuai, Hujun Bao,
  and Xiaowei Zhou.
\newblock Neural body: Implicit neural representations with structured latent
  codes for novel view synthesis of dynamic humans.
\newblock {\em 2021 IEEE/CVF Conference on Computer Vision and Pattern
  Recognition (CVPR)}, pages 9050--9059, 2021.

\bibitem{Pumarola2021DNeRFNR}
Albert Pumarola, Enric Corona, Gerard Pons-Moll, and Francesc Moreno-Noguer.
\newblock D-nerf: Neural radiance fields for dynamic scenes.
\newblock {\em 2021 IEEE/CVF Conference on Computer Vision and Pattern
  Recognition (CVPR)}, pages 10313--10322, 2021.

\bibitem{Rebain2021DeRFDR}
Daniel Rebain, Wei Jiang, Soroosh Yazdani, Ke Li, Kwang~Moo Yi, and Andrea
  Tagliasacchi.
\newblock Derf: Decomposed radiance fields.
\newblock {\em 2021 IEEE/CVF Conference on Computer Vision and Pattern
  Recognition (CVPR)}, pages 14148--14156, 2021.

\bibitem{Rombach2022HighResolutionIS}
Robin Rombach, A. Blattmann, Dominik Lorenz, Patrick Esser, and Bj{\"o}rn
  Ommer.
\newblock High-resolution image synthesis with latent diffusion models.
\newblock {\em 2022 IEEE/CVF Conference on Computer Vision and Pattern
  Recognition (CVPR)}, pages 10674--10685, 2022.

\bibitem{Schwarz2020GRAFGR}
Katja Schwarz, Yiyi Liao, Michael Niemeyer, and Andreas Geiger.
\newblock Graf: Generative radiance fields for 3d-aware image synthesis.
\newblock {\em ArXiv}, abs/2007.02442, 2020.

\bibitem{Shysheya2019TexturedNA}
Aliaksandra Shysheya, Egor Zakharov, Kara-Ali Aliev, Renat Bashirov, Egor
  Burkov, Karim Iskakov, Aleksei Ivakhnenko, Yury Malkov, I. Pasechnik, Dmitry
  Ulyanov, Alexander Vakhitov, and Victor~S. Lempitsky.
\newblock Textured neural avatars.
\newblock {\em 2019 IEEE/CVF Conference on Computer Vision and Pattern
  Recognition (CVPR)}, pages 2382--2392, 2019.

\bibitem{Sitzmann2019DeepVoxelsLP}
Vincent Sitzmann, Justus Thies, Felix Heide, Matthias Nie{\ss}ner, Gordon
  Wetzstein, and Michael Zollh{\"o}fer.
\newblock Deepvoxels: Learning persistent 3d feature embeddings.
\newblock {\em 2019 IEEE/CVF Conference on Computer Vision and Pattern
  Recognition (CVPR)}, pages 2432--2441, 2019.

\bibitem{Sorgi2011TwoViewGE}
Lorenzo Sorgi.
\newblock Two-view geometry estimation using the rodrigues rotation formula.
\newblock {\em 2011 18th IEEE International Conference on Image Processing},
  pages 1009--1012, 2011.

\bibitem{Srinivasan2021NeRVNR}
Pratul~P. Srinivasan, Boyang Deng, Xiuming Zhang, Matthew Tancik, Ben
  Mildenhall, and Jonathan~T. Barron.
\newblock Nerv: Neural reflectance and visibility fields for relighting and
  view synthesis.
\newblock {\em 2021 IEEE/CVF Conference on Computer Vision and Pattern
  Recognition (CVPR)}, pages 7491--7500, 2021.

\bibitem{Thies2019DeferredNR}
Justus Thies, Michael Zollh{\"o}fer, and Matthias Nie{\ss}ner.
\newblock Deferred neural rendering: Image synthesis using neural textures.
\newblock {\em arXiv: Computer Vision and Pattern Recognition}, 2019.

\bibitem{Tretschk2021NonRigidNR}
Edgar Tretschk, Ayush Tewari, Vladislav Golyanik, Michael Zollh{\"o}fer,
  Christoph Lassner, and Christian Theobalt.
\newblock Non-rigid neural radiance fields: Reconstruction and novel view
  synthesis of a dynamic scene from monocular video.
\newblock {\em 2021 IEEE/CVF International Conference on Computer Vision
  (ICCV)}, pages 12939--12950, 2021.

\bibitem{Weng2020Vid2ActorFA}
Chung-Yi Weng, Brian Curless, and Ira Kemelmacher-Shlizerman.
\newblock Vid2actor: Free-viewpoint animatable person synthesis from video in
  the wild.
\newblock {\em ArXiv}, abs/2012.12884, 2020.

\bibitem{Weng2022HumanNeRFFR}
Chung-Yi Weng, Brian Curless, Pratul~P. Srinivasan, Jonathan~T. Barron, and Ira
  Kemelmacher-Shlizerman.
\newblock Humannerf: Free-viewpoint rendering of moving people from monocular
  video.
\newblock {\em 2022 IEEE/CVF Conference on Computer Vision and Pattern
  Recognition (CVPR)}, pages 16189--16199, 2022.

\bibitem{Wu2020MultiViewNH}
Minye Wu, Yuehao Wang, Qiang Hu, and Jingyi Yu.
\newblock Multi-view neural human rendering.
\newblock {\em 2020 IEEE/CVF Conference on Computer Vision and Pattern
  Recognition (CVPR)}, pages 1679--1688, 2020.

\bibitem{Xian2021SpacetimeNI}
Wenqi Xian, Jia-Bin Huang, Johannes Kopf, and Changil Kim.
\newblock Space-time neural irradiance fields for free-viewpoint video.
\newblock {\em 2021 IEEE/CVF Conference on Computer Vision and Pattern
  Recognition (CVPR)}, pages 9416--9426, 2021.

\bibitem{Xu2021HNeRFNR}
Hongyi Xu, Thiemo Alldieck, and Cristian Sminchisescu.
\newblock H-nerf: Neural radiance fields for rendering and temporal
  reconstruction of humans in motion.
\newblock In {\em NeurIPS}, 2021.

\bibitem{Zhang2021NeRFactorNF}
Xiuming Zhang, Pratul~P. Srinivasan, Boyang Deng, Paul~E. Debevec, William~T.
  Freeman, and Jonathan~T. Barron.
\newblock Nerfactor: Neural factorization of shape and reflectance under an
  unknown illumination.
\newblock {\em ACM Trans. Graph.}, 40:237:1--237:18, 2021.

\end{thebibliography}
}

\end{document}